\documentclass[acmlarge]{acmart}
\AtBeginDocument{%
  \providecommand\BibTeX{{%
    \normalfont B\kern-0.5em{\scshape i\kern-0.25em b}\kern-0.8em\TeX}}}

\acmConference[]{}{}{}
\acmBooktitle{} 
\acmPrice{}
\acmISBN{}

\acmSubmissionID{}

\begin{document}

\title{Using Augmented Face Images to Improve Facial Recognition Tasks}

\author{Shuo Cheng}
\email{shuo.cheng@bytedance.com}
\author{Guoxian Song}
\email{guoxian001@e.ntu.edu.sg}
\author{Wan-Chun Ma}
\email{wanchun.ma@bytedance.com}
\author{Chao Wang}
\email{chao.wang@bytedance.com}
\author{Linjie Luo}
\email{linjie.luo@bytedance.com}

\affiliation{
  \\
  \institution{ByteDance Inc}
  \state{California}
  \city{San Jose}
  \country{USA}
}


\begin{abstract}
We present a framework that uses GAN-augmented images to complement certain specific attributes, usually underrepresented, for  machine learning model training. This allows us to improve inference quality over those attributes for the facial recognition tasks.
\end{abstract}




\maketitle

\section{Introduction}

The process of training a machine learning (ML) model is primarily about tuning its parameters such that it can map any input to a specific output. The number of samples that is required for training a model is usually proportional to the network capacity, the number of output dimensions, and most of all, the complexity of the task. For tasks that requires to operate in an arbitrary, in-the-wild environment, the amount of the data could be staggeringly large. For example, the effort to naively capture and label a sufficient dataset to cover all possible conditions for a face-related recognition task that predicts ages, races, hair styles, facial hairs, accessories (eyewear, earrings), illuminations, etc., could simply grow exponentially.

As the visual effects and video game industries driving the realism of computer generated imagery (CGI), we also begin to see the trend of using legacy CGI techniques to generate synthetic dataset for ML model training. Wood et al.~\cite{wood2021microsoftfakeit} has shown that it is possible to train neural nets for in-the-wild facial recognition tasks using only synthetically-generated images. Some of the benefits of having a synthetic dataset generation system are: 1) perfect labels can be obtained for free, since they are merely the control variables or derived properties for the aforementioned parametric model, and 2) we can generate samples as much as we want. Nevertheless, to create such a generation system, which may consists of a controllable face and body generators with a comprehensive library of 3D assets for components such as hairs and accessories, can still be a daunting work.

Something between a real-captured approach versus pure-synthesized approach is so-called augmentation. For example, typical augmentation operations for image data include geometric or color augmentations, random erasing, feature space augmentation, etc. More recently, StyleGAN~\cite{karras2019stylebased} and its variants (e.g.~\cite{Karras2019stylegan2}) are able to generate high quality and very natural looking synthetic face images. These techniques are heavily used for image augmentation as well, mostly for image transfer~\cite{CycleGAN2017, pix2pix2017} and stylization~\cite{Song:2021:AgileGAN}. InterFaceGAN~\cite{shen2020interpreting} also provides a way to identify semantic latent vectors in StyleGAN for further manipulation use. In this paper, we present a new framework that incorporates StyleGAN manipulation to generate augmented images for ML training. This strategy enhances the inference quality of facial recognition tasks. We will use the task that predicts facial blendshape activation from a facial image as example.

\section{Method}
\begin{figure}
\includegraphics[width=0.8\linewidth]{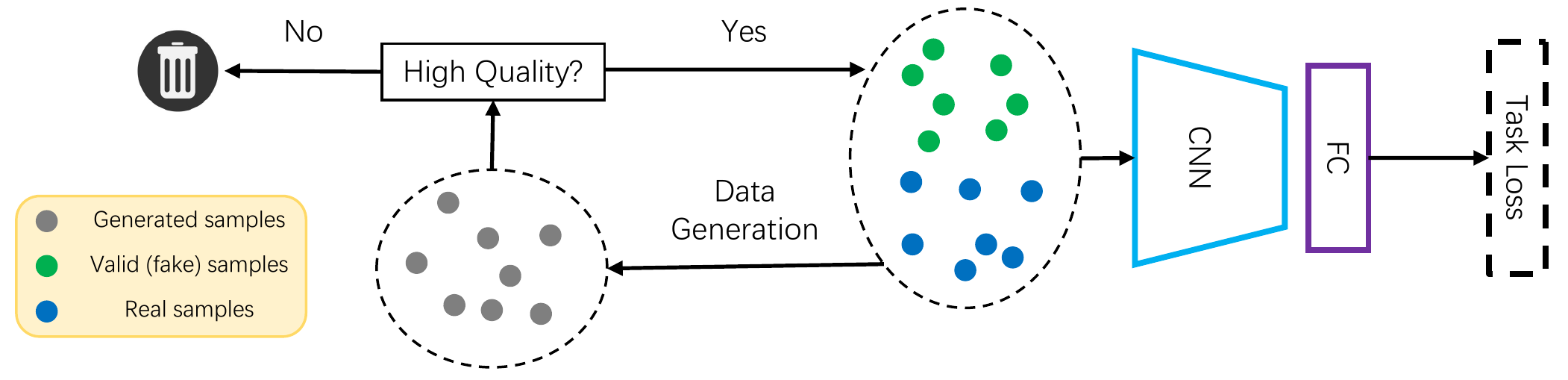}
\caption {Our workflow that uses generated high quality augmented samples to boost model performance.} 
\label{fig:workflow}
\end{figure}

A task that predicts facial blendshape activation (e.g. jawOpen, eyeBlink, etc.) from a facial image is commonly used for facial animation applications~\cite{10.1145/2010324.1964972, 10.1145/2980179.2980252}. A blendshape model is a linear deformable model based on a set of pre-defined shapes which represent the surface movement of various facial regions, and the "activation" is represented as the blending weights of the linear model~\cite{10.2312:egst.20141042}. To train a neural net that can perform such task, usually we prepare a paired dataset where each sample consists of a facial image and its corresponding blendshape activation, and the neural net can be trained in a supervised way. In order to diversify our training samples, our goal is to add features like eyewear or beards to the real-captured image, but also make sure that this operation does not alter the facial expression in that image, since it is tied to the paired activation. There are two major stages involved in our pipeline. First, we encode the input image to latent space, edit the latent codes and use a pre-trained StyleGAN2 to generate the final output image from the modified latent code. Second, After finishing the data augmentation, we use the off-the-shield landmark detector to extract the good samples from the augmentation. And finally, we can use the augmented images to train the network to boost its performance (as shown in Fig.~\ref{fig:workflow}).

\subsection{Data Augmentation}
For data augmentation, we follow the InterFaceGAN~\cite{shen2020interpreting} method to identify those semantic latent vectors in StyleGAN. We use a pre-trained StyleGAN encoder~\cite{tov2021designing} to map the images inversely into latent space with some attributes labels, and then apply an independent linear SVMs to each attribute. After that, we can find a hyperplane of each SVMs and its corresponding normal vector which can be associated with semantic meaning. To augment each image in the dataset, we first encode each image into the latent codes, then add a randomly-scaled semantic vector to the codes. This should promote the diversity of augmentation. At last, the augmented codes are forwarded into a pre-trained StyleGAN2 to synthesize the final output image.

\subsection{Learning with Generated Samples}
Our goal is create training images as much as possible that can re-use the original output (blendshape activation). However, if the augmented image contains undesired facial expression changes, which violates the underlying blendshape activation, then this could jeopardize the inference quality. Specifically, for each input image $x \in R^{H\times W\times3}$, we want the generated image $\hat{x} \in R^{H\times W\times3}$ keeps unchanged on task-related attributes but except the modified attributes (i.e., beard, glasses). To measure the validity of the generated samples, we first utilize off-the-shelf face key-point detectors~\cite{sun2013deep, zhang2014facial} for extracting face landmarks: $\Phi: R^{H\times W\times3} \rightarrow R^{240\times2}$, and then we compute the normalized quality score as the root-mean-square error between the landmarks from the original image and its generated counterpart. We extract the generated images that have quality score above the given threshold. With the mixed dataset containing the generated images with our intended visual attributes, we can then train a model that is more robust to those underrepresented attributes. For the implementation, we use a convolutional neural network~\cite{lecun1995convolutional} as feature extractor and a small number of fully-connected layers for regressing task-related values. The smooth L1 loss is used and the network is trained with the Adam optimizer~\cite{kingma2014adam}.

\section{Results and Conclusion}
\begin{figure*}
	\centering
	\includegraphics[width=0.8\linewidth]{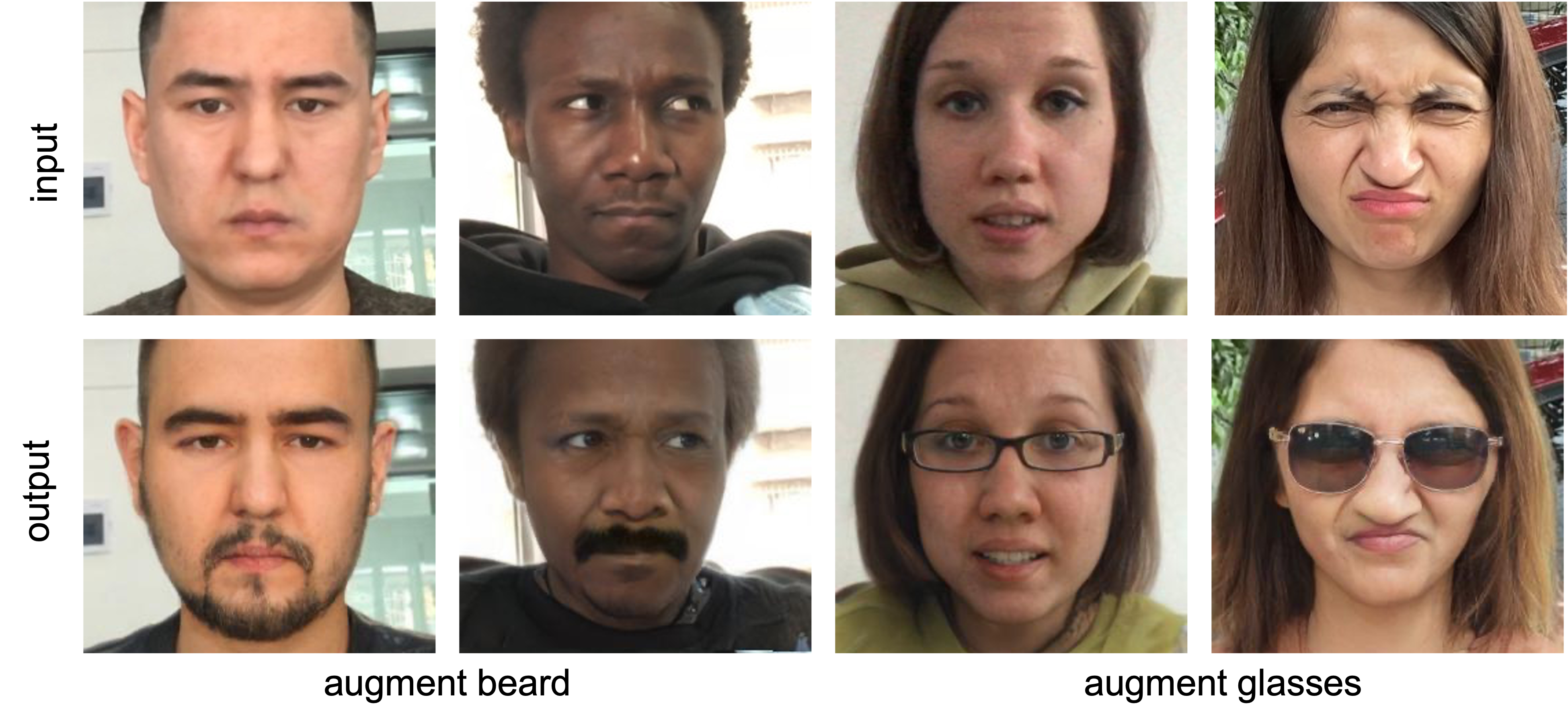}
	\vspace{-10pt}
	\caption[Caption for LOF]{Augment input images attributes while maintain the expressions.} 
	\label{fig:styles_real}
\end{figure*}

\begin{figure*}
	\centering
	\includegraphics[width=0.8\linewidth]{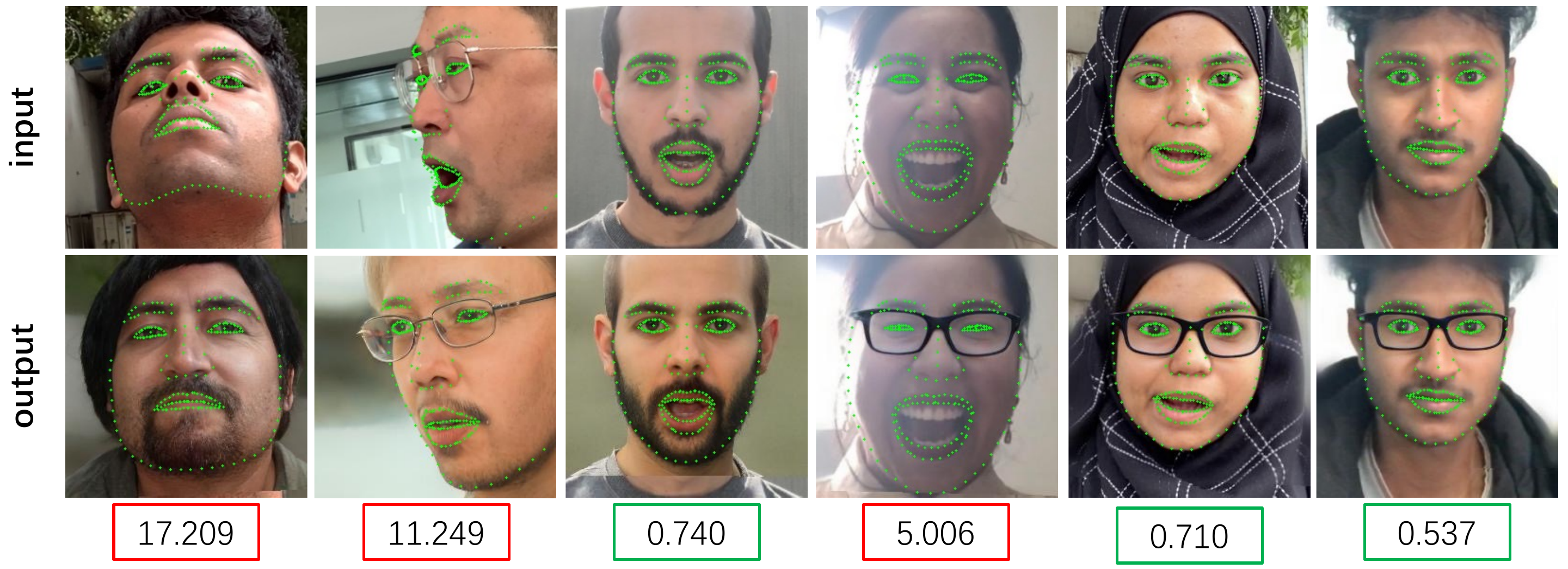}
	\vspace{-10pt}
	\caption[Caption for LOF]{Automatic quality assessment based on landmark disagreement. The images with error showing in red rectangle are not used.} 
	\label{fig:data_filter}
\end{figure*}

\begin{figure*}
	\centering
	\includegraphics[width=0.8\linewidth]{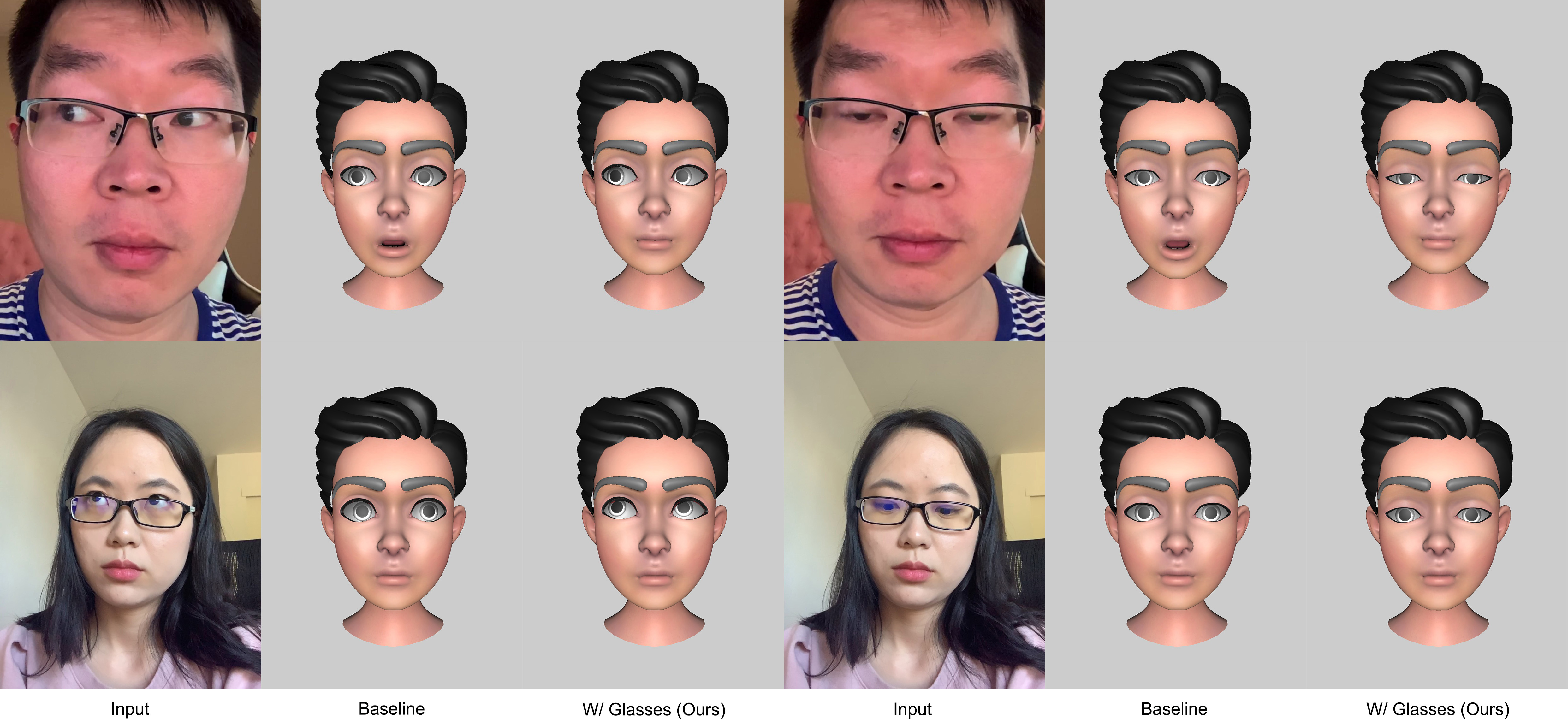}
	\vspace{-10pt}
	\caption[Caption for LOF]{Visual comparison of different methods on recognizing facial blendshape activation.} 
	\label{fig:qualitative}
\end{figure*}

We visualize some of the generated images with selected attributes in Fig.~\ref{fig:styles_real}, the results demonstrate that our approach can generate realistic images with the modified attributes but keep the expression intact. Fig.~\ref{fig:data_filter} showing that the proposed landmark disagreement metric serves as a cue for filtering out the low quality samples that are not suitable for training.
To verify the effectiveness of training with the generated data, we test our learned models with different training sources. The quantitative results on test benchmark are reported with data across different expressions and emotions such as blinking eyes, speaking, happy, surprise, etc. The error metrics include L1 mean, L1 minimum, and L1 maximum. Please see Table~\ref{benchmark} for more details. We also present qualitative examples in Fig.~\ref{fig:qualitative}. We found that learning with the augmented data will boost the performance, and filter the low quality data with some heuristic rules (i.e., landmark errors) can further improve the model performance.

\begin{table}[]
\begin{tabular}{ccccc}
\hline
Methods                                    & Blinking Eyes                          & Rolling Eyes                           & Speaking                               & All                                   \\ \hline
\multicolumn{1}{c|}{Baseline}              & \multicolumn{1}{c|}{0.058/0.031/0.107} & \multicolumn{1}{c|}{0.070/0.030/0.127} & \multicolumn{1}{c|}{0.059/0.031/0.102} & 0.064/0.032/0.112                     \\
\multicolumn{1}{c|}{W/ Beard}              & \multicolumn{1}{c|}{0.054/0.033/0.085} & \multicolumn{1}{c|}{0.066/0.028/0.120} & \multicolumn{1}{c|}{0.057/0.033/0.087} & 0.060/0.033/0.100                     \\
\multicolumn{1}{c|}{W/ Glasses}            & \multicolumn{1}{c|}{0.049/0.032/0.080} & \multicolumn{1}{c|}{0.058/0.027/0.101} & \multicolumn{1}{c|}{0.052/0.034/0.088} & 0.058/0.033/0.094                     \\
\multicolumn{1}{l|}{W/ Beard (filtered)}   & \multicolumn{1}{l|}{0.048/0.031/0.083} & \multicolumn{1}{l|}{0.054/0.027/0.098} & \multicolumn{1}{l|}{0.052/0.030/0.088} & \multicolumn{1}{l}{0.056/0.031/0.093} \\
\multicolumn{1}{l|}{W/ Glasses (filtered)} & \multicolumn{1}{l|}{0.050/0.033/0.083}  & \multicolumn{1}{l|}{0.059/0.032/0.096} & \multicolumn{1}{l|}{0.050/0.033/0.079}  & \multicolumn{1}{l}{0.055/0.032/0.089} \\ \hline
\end{tabular}
\caption{We report the performance of our learned models with different training sources, the errors metrics are L1 mean/L1 minimum/L1 maximum, and the lower the better. The performance are measured on our test benchmark with data across different expressions and emotions such as blinking eyes, speaking, happy, surprise, etc.}
\label{benchmark}
\end{table}

\bibliographystyle{ACM-Reference-Format}
\bibliography{sample-acmtog}

\end{document}